# Recognition of Isolated Words using Zernike and MFCC features for Audio Visual Speech Recognition

Prashant Borde[a]*, Amarsinh Varpe[b], Ramesh Manza[c], Pravin Yannawar[a]*,

[ab]*Vision and Intelligent System Lab,*
[c]*Biomedical Image Processing Lab,*
*Department of Computer Science and IT,*
*Dr. Babasaheb Ambedkar Marathwada University, Aurangabad (MS) India.*

**Abstract**

Automatic Speech Recognition (ASR) by machine is an attractive research topic in signal processing domain and has attracted many researchers to contribute in this area. In recent year, there have been many advances in automatic speech reading system with the inclusion of audio and visual speech features to recognize words under noisy conditions. The objective of audio-visual speech recognition system is to improve recognition accuracy. In this paper we computed visual features using *Zernike moments* and audio feature using *Mel Frequency Cepstral Coefficients (MFCC)* on '*vVISWa*' *(Visual Vocabulary of Independent Standard Words)* dataset which contains collection of isolated set of city names of 10 speakers. The visual features were normalized and dimension of features set was reduced by Principal Component Analysis (PCA) in order to recognize the isolated word utterance on PCA space.The performance of recognition of isolated words based on visual only and audio only features results in 63.88% and 100% respectively.

## 1. Introduction

In the recent year, there are many automatic speech reading system proposed that combine audio as well as visual speech features. In computer speech recognition visual component of speech is used for support of acoustic speech recognition. Design of an audio-visual speech recognizer is based on human lip-reading expert experiences. Hearing impaired people achieve recognition rate of 60-80% in dependence on lip-reading conditions. Most important conditions for good lip-reading are quality of visual speech of a speaker (proper articulation) and angle of view. Sometimes people, who are well understood from acoustic component may be not well lip-read but for hearing-impaired or even deaf people visual speech component is important source of information. Lip-reading (visual speech recognition) is used by people without disabilities, too. It helps better understanding in case when the acoustic speech is less intelligible.Task of automatic speech recognition by a computer, when both acoustic and visual component of speech is used has attracted many researcher to contribute in automatic*Audio-Visual Speech Recognition* domain. This is challenging because of the visual articulations vary with speaker to speaker and can contain very less information as compared to acoustic signal therefore identification of robust features is still center of attraction of many researchers.The visual speech component is used by hearing-impaired people (for lip-reading), but is also used unconsciously by all people in common communication, especially in noisy environment by ŽELEZNÝ*et.al*,2006.

* Corresponding author. Tel.: +912402403316; fax: +912402403317.
  *E-mail address:* borde.prashantkumar@gmail.com, pravinyannawar@gmail.com.

J R Deller*et al*.,1993, has described in his work on Automatic speech recognition (ASR) for well define applications like dictations and medium vocabulary transaction processing tasks are in relatively controlled environments have been designed. It is observed by the researchers that, the ASR performance was far from human performance in variety of tasks and conditions, indeed ASR to date is very sensitive to variations in the environmental channel (non-stationary noise sources such as speech babbled, reverberation in closed spaces such as car, multi-speaker environments) and style of speech (such as whispered etc). Eric Petjan*et al.,* 1987 and Finn K.I *et al.,* 1986 worked on Lip-reading is an auditory, imagery system as a source of speech and image information. It provides the redundancy with the acoustic speech signal but is less variable than acoustic signals; the acoustic signal depends on lip, teeth, and tongue position to the extent that significant phonetic information was obtainable using lip movement recognition alone. The intimate relation between the audio and imagery sensor domains in human recognition can be demonstrated with McGurk Effect by McGurk*et al.,* 1976 and 1978.where the perceiver "hears" something other than what was said acoustically due to the influence ofconflicting visual stimulus. The current speech recognition technology may perform adequately in the absence of acoustic noise for moderate size vocabularies; but even in the presence of moderate noise it fails except for very small vocabularies by Paul D.B *et al*., Malkin F.J, 1986, Meisel W.S, 1987 and Moody T *et al*., 1987.

Humans have difficulty distinguishing between some consonants when acoustic signal is degraded. However, to date all automatic speech reading studies have been limited to very small vocabulary tasks and inmost of cases to very small number of speakers. In addition the numbers of diverse algorithms have been suggested in the literature for automatic speech reading and areverydifficult to compare, as they are hardly ever tested on any common audio visual databases. Furthermore, most of such databases are very small duration thus placing doubts about generalization of reported results to large population and tasks. There is no specific answer to this but researchers are concentrating more on speaker independent audio-visual large vocabulary continuous speech recognition systems by ChalapathyNeti, et.al., 2000.

Many methods have been proposed by researchers in order to enhance speech recognition system by synchronization of visual information with the speech as improvement on automatic Lip-reading system which incorporates dynamic time warping, and vector quantization method applied on alphabets, digits and The recognition was restricted to isolated utterances and was speaker dependent by Eric Petjan et al.,1987. Later ChristophBregler.,1993had worked on how recognition performance in automated speech perception can be significantly improved & introduced an extension to existing Multi- State Time Delayed Neural Network architecture for handling both the modalities that is acoustics and visual sensor input. Similar work have been done by Yuhaset.al.,1993& focused on neural network for vowel recognition and worked on static images. Paul Duchnowskiet.al.,1995 worked on movement invariant automatic Lip-reading and speech recognition, JuergenLuettin., 1996 used active shape model and hidden markov model for visual speech recognition, *K.L.* Sum et al.,2001 proposed a new optimization procedure for extracting the point-based lip contour usingactive shape model, Capiler 2001 used Active shape model and Kalman filtering in spatiotemporal for noting visual deformations, Ian Matthews et al.,2002 has proposed method for extraction of visual features of Lip-reading for audio-visual speech recognition , Xiaopeng Hong et al.,2006 used PCA based DCT features Extraction method for Lip-reading, Takeshi Saitohetal., 2008 has analyzed efficient Lip-reading method for various languages where they focused on limited set of words from English, Japanese, Nepalese, Chinese, Mongolian. The words in English and their translated words in above listed languages were considered for the experiment.Meng Li et al.,2008 has proposed A Novel Motion Based Lip Feature Extraction for Lip-reading problems.

This paper introduced mechanism of extraction of audio-visual features for visual speech recognition. The content of the paper is organized in five section, section II deals with *'vVISWa'* dataset, Section III deals with methodology adopted, section IV deal with experimental results, section V is conclusion of work followed acknowledgement and references.

## 2. *'vVISWa'* Dataset

Many researchers have defined their own dataset and very few are available online freely. Indeed it is very difficult to distribute the data base freely on the web due to the size. The video sequences used for this study was collected in the laboratory in a closed environment. The 'vVISWa' (*Visual Vocabulary of Independent Standard Words*) database consists set of independent/isolated standard words from Marathi, Hindi and English script. The dataset of isolated city words like {*'Aurangabad', 'Beed', 'Hingoli', 'Jalgaon', 'Kolhapur', 'Latur', 'Mumbai', 'Osmanabad', 'Parbhani', 'Pune', 'Satara', 'Solapur'*} in M*arathi* were considered for this experiments. Each visual utterance is recorded for 2 second. The database consists of 10 individuals speakers, 4 male and 6 female and each speaker speaking each word utterance for 10 times. Each individual has uttered word in close-open-close constraint without head movement. The database comprised of 1200 utterance (10*10*12) of these independent standard words. The figure 2.1 shows the experimental arrangement of acquisition of utterances from individual speaker.

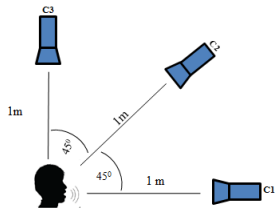

Fig. 2.1 Acquisition of utterances

The visual utterance was recorded at resolution of 720 x 576 in (*.avi) format using high definition digital camera in three angle comprising full frontal using camera 'C1', 45ºface using camera 'C2' and side pose using camera 'C3'. This study was focused on full frontal profile of utterance. Lighting was controlled and a dark gray background was used. The recognition of visual utterance of word, the data set of isolated city words was divided in to 70% known sample and 30% unknown sample. The System was trained over 70% known set and 30% unknown samples were tested on the known set and success recognition of isolated word based on visual utterance was evaluated.

## 3. Methodology

The typical audio visual speech recognition system accepts the audio and visual input as shown in figure 3.1. The audio input is captured with the help of standard audio mic and visual utterance is captured by using standard camera. The place between camera and individual speaker is kept constant in order to get proper visual utterance. Once the input is acquired, it will be preprocessed for acoustic feature extraction and visual feature extraction separately and further used for recognition and integration of utterance.

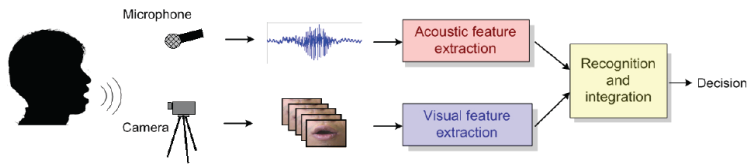

**Fig. 3.1** Organization of AVSR

Visual features can be grouped into three general categories: shape-based, appearance-based, and combinational approaches. All three types require the localization and tracking of Region of Interest (ROI). Region of interest for computation of visual feature will be concentrated towards the movement of lips (opening and closing of mouth) over the time frame which is very complex. In-view of this calculating good and discriminatory visual feature of mouth plays vital role in the recognition.

*3.1 Region-of-Interest (ROI) Detection / Localization*

The visual information relevant to speech is mostly contained in the motion of visible articulators such as lips, tongue and jaw. In order to extract this information from a sequence of video frames it is advantageous to track the complete motion of the face detectionand mouth localization, this helps in visual feature extraction. In order to achieve robust and real-time face detection we used the '*Viola-Jones*' detector based on '*AdaBoost*' which is a binary classifier that uses cascades of weak classifiers to boost its performance by G. Bradski and A. Kaehler, 2008 and C. M. Bishop,2006. In our study we used detector to detect the face in each frame from the sequence and subsequently mouth portion of the face is detected. This is achieved by finding the median of the coordinates of the ROI object *bounding box* of frames. Finally a region-of-interest (ROI)is extracted by resizing the mouth bounding box to120x120 pixels size as shown in figure3.2.

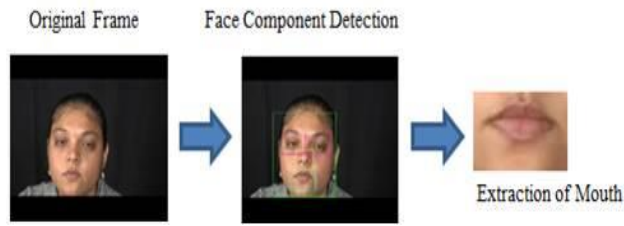

**Fig. 3.2** Mouth Localization using Viola-Jones Algorithm

After isolating ROI Mouth frame, each frame from utterance was pre-processed so as to obtain good discriminating features. The preprocessing include, separation of RGB channel and subtract R channel to gray scale image, filter the resultant image then calculating gray threshold range and converted frame into binary frame to get the actual Region-of-Interest (ROI)of containing only the portion covered by lips as shown in figure 3.3 and in similar way the ROI for each frame of utterance is identified as shown Table 1

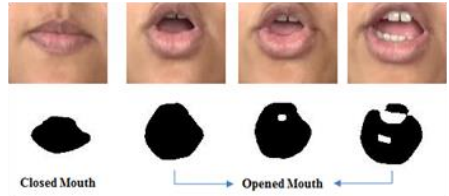

**Fig. 3.3** Isolation of ROI Identification

| No. of Frame | Frame 1 | Frame 2 | Frame 3 | Frame 4 | Frame 5 | Frame 6 | Frame 7 | Frame 8 | … | Frame 52 |
|---|---|---|---|---|---|---|---|---|---|---|
| Processed Frame | | | | | | | | | … | |

**Table 1:** Isolated Mouth from video sequence

### 3.2 Visual Feature Extraction

After extracting the region-of-interest (ROI) from frame, the next step is to extract the appropriate features in each frame. We used *Zernike movement* for features extraction. *Zernike moments* have mathematical properties which makes them ideal for image feature extraction. These moments are best suited for describing shape and shape classification problem. *Zernike moments*, a type of moment function, are the mapping of an image onto a set of complex Zernike polynomials which form a complete orthogonal set on the unit disk with $(x^2 + y^2) = 1$.

$$Z_{mn} = \frac{m+1}{\pi} \int\int_{x\ y} I(x,y)[V_{mn}(x,y)]\,dxdy \qquad (1)$$

Where '*m*' defines the order of Zernike polynomial of degree '*m*', '*n*' defines the angular dependency, and $I(x,y)$ be the gray level of a pixel of video frame on which the moment is calculated. The Zernike polynomials $V_{mn}(x,y)$ are expected in polar coordinates using radial polynomial ($R_{mn}$) as per equation (2) and (3)

$$V_{mn}(r,\theta) = R_{mn}(r)e^{-jn\theta} \qquad (2)$$

$$R_{mn}(r) = \sum_{s=0}^{\frac{m-|n|}{2}} (-1)^s \frac{(m-s)!}{s!\left[\frac{m+|n|}{2}-s\right]!\left[\frac{m-|n|}{2}-s\right]!} r^{m-2s} \qquad (3)$$

The resultant *Zernike moments* $Z_{mn}$ are invariant under *rotation*, *scale and translational* changes. Zernike Moments are the pure statistical measure of pixel distribution around centre of gravity of shape and allows capturing information just at single boundary point. They can capture some of the global properties missing from the pure boundary-based representations like the overall image orientation was proposed by JyotsnaraniTripathy,*2010* and Sun-Kyoo Hwang et al., 2006. Table 2 shows calculated Zernike moment feature up to 9$^{th}$ order for a sample frame from sequence as

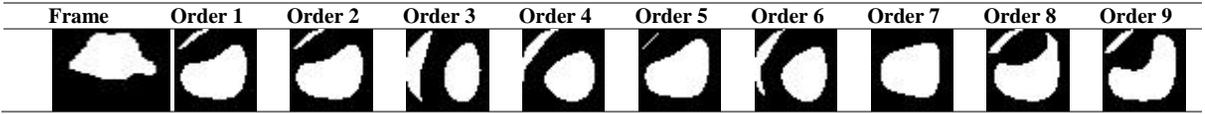

| Frame | Order 1 | Order 2 | Order 3 | Order 4 | Order 5 | Order 6 | Order 7 | Order 8 | Order 9 |

**Table 2:** Computed Zernike moment of 9$^{th}$ order for i$^{th}$ frame from video sequence.

These feature were calculated for all the samples of '*Training set*' and '*Test set*'. The '*Zernike Feature Matrix*' for the samples of '*Training set*', are as shown in table 2.a) and '*Test set*' are as shown in 2.b)

| Known Set Frame No | Moment 1 | Moment 2 | Moment 3 | Moment 4 | Moment 5 | Moment 6 | Moment 7 | Moment 8 | Moment 9 |
|---|---|---|---|---|---|---|---|---|---|
| 1 | 113.6623 | 83.8783 | 97.2958 | 22.79026 | 93.09135 | 20.30108 | 17.2095 | 34.9933 | 38.6219 |
| 2 | 130.2606 | 81.1683 | 107.603 | 9.117232 | 82.55588 | 17.8117 | 40.1537 | 7.79399 | 34.6609 |
| 3 | 123.5299 | 60.8899 | 76.2613 | 6.61097 | 53.6311 | 12.62223 | 86.0334 | 20.0651 | 37.212 |
| 4 | 96.47357 | 52.5382 | 50.5424 | 12.6747 | 43.6893 | 18.11593 | 77.4625 | 31.40335 | 31.1274 |
| 5 | 133.6902 | 65.0204 | 97.7663 | 2.88031 | 55.25426 | 15.06808 | 60.9075 | 14.38177 | 42.9087 |
| 6 | 137.4431 | 82.1828 | 105.348 | 7.687009 | 82.90935 | 24.91766 | 40.698 | 5.5325 | 49.5998 |
| 7 | 141.7249 | 88.4639 | 98.0411 | 8.199153 | 81.44456 | 29.86014 | 57.4904 | 6.10306 | 58.5721 |
| 8 | 137.8847 | 81.6338 | 80.8799 | 3.86494 | 69.05018 | 34.27437 | 77.0594 | 15.43476 | 59.1316 |
| 9 | 132.6377 | 87.2031 | 118.214 | 20.16833 | 91.06382 | 21.75958 | 20.4497 | 27.6913 | 41.8429 |
| ⋮ | ⋮ | ⋮ | ⋮ | ⋮ | ⋮ | ⋮ | ⋮ | ⋮ | ⋮ |

**Table 2.a)** Zernike moment features for training set

| Unknown Set Frame No | Moment 1 | Moment 2 | Moment 3 | Moment 4 | Moment 5 | Moment 6 | Moment 7 | Moment 8 | Moment 9 |
|---|---|---|---|---|---|---|---|---|---|
| 1 | 121.5122 | 70.6177 | 88.3392 | 0.58559 | 62.22295 | 20.91341 | 55.2083 | 12.17358 | 37.7686 |
| 2 | 118.0057 | 57.4738 | 75.4854 | 8.49015 | 51.32621 | 14.75032 | 70.3078 | 23.14418 | 39.8614 |
| 3 | 121.3839 | 69.8241 | 81.3716 | 0.65449 | 65.21946 | 17.72637 | 61.4222 | 10.05763 | 40.8054 |
| 4 | 136.9246 | 85.9918 | 89.025 | 0.74707 | 84.20651 | 31.30934 | 64.9828 | 8.079067 | 55.197 |
| 5 | 169.4281 | 81.0754 | 134.969 | 9.86071 | 80.96385 | 19.45107 | 42.3216 | 13.4204 | 64.8695 |
| 6 | 127.3445 | 78.3759 | 66.6737 | 2.870781 | 79.1795 | 11.6277 | 94.4969 | 4.71345 | 46.6652 |
| ⋮ | ⋮ | ⋮ | ⋮ | ⋮ | ⋮ | ⋮ | ⋮ | ⋮ | ⋮ |

**Table 2.b)** Zernike movement features for Test Set

## 3.3 Principal Component Analysis

Zernike features for each frame in visual utterance was computed and results in to 9x1 columns. The visual utterance was captured for two seconds and results in formation of 52 frames therefore the Zernike features for one visual utterance results in to 468x1 for single word. Similarly all visual words of city names from *'vVISWa'* are passed for visual feature extraction which results in 468x72 size matrix. This feature set was called as *'Training Set'* and the dimension of this data set is to be reduced and to be interpreted using Principal Component Analysis. PCA was applied on Zernike features to extract the most significant components of feature corresponding to the set of isolated words. PCA converts all of the original variables to be some independent linear set of variables. Those independent linear sets of variables possess the most information in the original data referred as principal components.

Following steps show the how to reduce the size of data and recognition using PCA.

> Step 1: Convert the all Zernike features into the column matrix as 'T'.
> Step 2: Calculate the *Mean* Column Vector 'm' for 'T'.
> Step 3: Computing the difference for each vector set $A_i = T_i - m$ where (i=1, 2 …N)
> Step 4: Calculating a *covariance* matrix $C = A*A'$
> Step 5: Calculate the *eigenvalues* and unit *eigenvectors* of the *covariance* Matrix 'C'.
> Step 6: Sort the *eigenvalues*.
> Step 7: Solve the mapping *eigenvectors* and project data on *Eigen space* for matching.

## 3.4 Acoustic Features Extraction using MFCC

The acoustic signals of *'vVISWa'* dataset corresponding to isolated words of city names have been processed for feature extraction. Recognition of uttered word is based on information in speech signal contained at the time of pronunciation of word. Mel-Frequency Cepstral Coefficients (MFCC) approach is the most popular because it uses spectral base as parameters for recognition. MFCC's are the coefficients, which represent audio based on perception of human auditory systems. The reason behind selection of MFCC for recognition purpose due to its peculiar difference between the operations of FFT/DCT. In the MFCC, the frequency bands are positioned logarithmically (on the Mel scale) which approximates the human auditory system's response more closely than the linearly spaced frequency bands of FFT or DCT by Clarence Goh Kok Leonet al., 2009 and B. Gold et al., 2000. Figure 3.4 shows the block diagram of MFCC features extraction process. After acquisition of input our objective is to remove noise contained in speech signal and it will be removed by using first order high pass filter. This process helps in cleaning acoustic-input. Frame blocking phase converts segmented concatenated voiced speech signals in to frames. The discontinuities contained in the segmented signal is checked at the beginning and end of each frame using hamming window and this was performed by windowing step. Fast Fourier Transform (FFT) brings the each frame of signal from time domain to frequency domain and the result is said to be 'spectrum'.

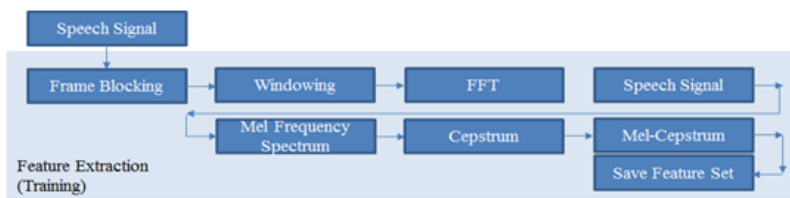

Figure 3.4 Block diagram of MFCC Feature extraction

It is well known that, human ear perception of frequency contents of sounds for speech signal does not follow a linear scale. Therefore, for each tone with an actual frequency f, measured in Hz, a subjective pitch is measured on a scale called the "*mel*" scale.The *mel* frequency scale is a linear frequency spacing below 1000 Hz and a logarithmic spacing above 1000Hz. To compute the *mels* for a given frequency *'f'* in Hz is calculated as.

$$\text{Mel}(f) = S_k = 2595 * \log_{10}(1 + \frac{f}{700}) \qquad (4)$$

Finally, we convert the calculated log *mel* spectrum back to time. The result is called the *Mel Frequency Cepstrum Coefficients (MFCC)*. The Cepstral representation of the speech signal spectrum provides a good representation of the local spectral properties of the signal forframe analysis. The *mel* spectrum coefficients are real numbers so we convert them to the time domain using the discrete cosine transform (DCT) and by this conversion contribution of pitch is removed by Vibha Tiwari.

$$C_n = \sum_{k=1}^{k}(\log S_k)\left[n\left(k - \frac{1}{2}\right)\frac{\pi}{k}\right] \qquad (5)$$
Where n=1,2,3...k

## 4. Experiment and Result

The recognition of visual only sample were carried out by Zernike moment and PCA. Similarly MFCC features were extracted for audio only recognition of uttered isolated words. These features were calculated for all sample of training set and stored for recognition purpose. The entire data set of isolated city names (visual only and audio only) were divided into 70-30 ratio that is 70% (Training samples) and 30% (Test samples). The recognition of isolated city name was divided in to two phase's first recognition of visual speech and second recognition of audio speech.

*4.1 Visual Speech Recognition*

Visual feature using Zernike feature were computed for all visual samples of training and test set. The training and test set were loaded for recognition of words at PCA space. This feature vector of training sample and test sample were checked for similarity using '*Euclidean*' distance classifier at PCA projection space. The 'Euclidean' distance provides information between each pair (one vector from test set and other vector from training set) of observations. The distance matrix of training sample and test sample using Zernike moment and PCA was shown in table 3.a)

It was seen that out of 36 samples 23 samples were correctly recognized and 13 samples were misclassified. Samples of '*Latur*' and '*Solapur*' were not recognized whereas samples of {'*Mumbai*', '*Osmanabad*', '*Parbhani*' and '*Satara*'} were recognized completely. It was seen that for three samples of 'Aurangabad' two samples were correctly recognized and one sample was mated with 'Kolhapur', this result in 66.66% of recognition of word. Similar results were observed for isolated word {'*Hingoli*', Jalna' ,'Kolhapur'and Pune'}. For isolated word '*Beed*' it was seen that out of three samples only one was correctly recognized and two samples were matched with word '*Pune*'.

9*4.2 Audio Speech Recognition*

MFCC features for all audio speech samples from training set and test set were computed. The '*Euclidean*' distance classifier was used for measuring the similarity between test samples and training sample. The table 3.b shows confusion matrix for acoustic speech recognition based on MFCC features. It was seen that all samples out of 36 samples were recognized correctly and over all recognition rate measured to be 100%.

Table 3.c provides the overall recognition rate for recognition of isolated words from '*vVISWa*' data set. This clearly indicates that the Zernike feature with PCA will helpful in recognition of words similar to the accuracy of recognition of words by hearing impaired peoples in dependence of lip-reading conditions. This approach will also be helpful in recognition of dictionary based words on visual only features when the acoustic signal is degraded due to noise.

| Method | Result |
|---|---|
| Visual Speech Recognition (Zernike + PCA) | 63.88 % |
| Acoustic Speech Recognition (MFCC) | 100% |

**Table 3.c)** Performance of Recognition of System

## 5. Conclusion

This paper, we described a complete Audio-Visual Speech Recognition system which include face and lip detection, features extraction and recognition. These experiment carried out using our own database, which was designed for evaluation purpose. We used a new technique for visual speech recognition that is Zernike Moment. The Zernike moment features are found to be more reliable and accurate feature for visual speech recognition and PCA used to reduce the dimension of data.

**Acknowledgements**

The Authors gratefully acknowledge support by the Department of Science and Technology (DST) for providing financial assistance for Major Research Project sanctioned under *Fast Track Scheme for Young Scientist*, vide sanction number SERB/1766/2013/14 and the authorities of Dr. Babasaheb Ambedkar Marathwada University, Aurangabad (MS) India, for providing the infrastructure for this research work.

| Words to Test | Visual Utterance | Training Sample | | | | | | | | | | | | Recognition | | |
|---|---|---|---|---|---|---|---|---|---|---|---|---|---|---|---|---|
| | | Aurangabad | Beed | Hingoli | Jalgaon | Kolhapur | Latur | Mumbai | Osmanabad | Perbhani | Pune | Satara | Solapur | Recognized | Miss | Accuracy |
| | | 1 | 2 | 3 | 4 | 5 | 6 | 7 | 8 | 9 | 10 | 11 | 12 | | | |
| 3 | Aurangabad | 2 | 0 | 0 | 0 | 1 | 0 | 0 | 0 | 0 | 0 | 0 | 0 | 2 | 1 | 66.66% |
| 3 | Beed | 0 | 1 | 0 | 0 | 0 | 0 | 0 | 0 | 0 | 2 | 0 | 0 | 1 | 2 | 33.33% |
| 3 | Hingoli | 1 | 0 | 2 | 0 | 0 | 0 | 0 | 0 | 0 | 0 | 0 | 0 | 2 | 1 | 66.66% |
| 3 | Jalgaon | 0 | 0 | 0 | 2 | 0 | 0 | 0 | 0 | 1 | 0 | 0 | 0 | 2 | 1 | 66.66% |
| 3 | Kolhapur | 0 | 0 | 0 | 0 | 2 | 1 | 0 | 0 | 0 | 0 | 0 | 0 | 2 | 1 | 66.66% |
| 3 | Latur | 0 | 0 | 0 | 0 | 0 | 0 | 0 | 0 | 0 | 0 | 0 | 0 | 0 | 3 | 0% |
| 3 | Mumbai | 0 | 0 | 0 | 0 | 0 | 0 | 3 | 0 | 0 | 0 | 0 | 0 | 3 | 0 | 100% |
| 3 | Osmanabad | 0 | 0 | 0 | 0 | 0 | 0 | 0 | 3 | 0 | 0 | 0 | 0 | 3 | 0 | 100% |
| 3 | Perbhani | 0 | 0 | 0 | 0 | 0 | 0 | 0 | 0 | 3 | 0 | 0 | 0 | 3 | 0 | 100% |
| 3 | Pune | 0 | 1 | 0 | 0 | 0 | 0 | 0 | 0 | 0 | 2 | 0 | 0 | 2 | 1 | 66.66% |
| 3 | Satara | 0 | 0 | 0 | 0 | 0 | 0 | 0 | 0 | 0 | 0 | 3 | 0 | 3 | 0 | 100% |
| 3 | Solapur | 0 | 0 | 0 | 0 | 0 | 0 | 0 | 0 | 0 | 0 | 0 | 0 | 0 | 3 | 0% |
| | | | | | | | | | | | | | Total | 23 | 13 | |
| | | | | | | | | | | | | | Final Result | 63.88% | | |

Table 3.a) Confusion Matrix for Visual Utterance Recognition using PCA based on Zernike Features

| Words to Test | Acoustic Utterance | Training Sample | | | | | | | | | | | | Recognition | | |
|---|---|---|---|---|---|---|---|---|---|---|---|---|---|---|---|---|
| | | Aurangabad | Beed | Hingoli | Jalgaon | Kolhapur | Latur | Mumbai | Osmanabad | Perbhani | Pune | Satara | Solapur | Recognized | Missed | Accuracy |
| | | 1 | 2 | 3 | 4 | 5 | 6 | 7 | 8 | 9 | 10 | 11 | 12 | | | |
| 3 | Aurangabad | 3 | 0 | 0 | 0 | 0 | 0 | 0 | 0 | 0 | 0 | 0 | 0 | 3 | 0 | 100% |
| 3 | Beed | 0 | 3 | 0 | 0 | 0 | 0 | 0 | 0 | 0 | 0 | 0 | 0 | 3 | 0 | 100% |
| 3 | Hingoli | 0 | 0 | 3 | 0 | 0 | 0 | 0 | 0 | 0 | 0 | 0 | 0 | 3 | 0 | 100% |
| 3 | Jalgaon | 0 | 0 | 0 | 3 | 0 | 0 | 0 | 0 | 0 | 0 | 0 | 0 | 3 | 0 | 100% |
| 3 | Kolhapur | 0 | 0 | 0 | 0 | 3 | 0 | 0 | 0 | 0 | 0 | 0 | 0 | 3 | 0 | 100% |
| 3 | Latur | 0 | 0 | 0 | 0 | 0 | 3 | 0 | 0 | 0 | 0 | 0 | 0 | 3 | 0 | 100% |
| 3 | Mumbai | 0 | 0 | 0 | 0 | 0 | 0 | 3 | 0 | 0 | 0 | 0 | 0 | 3 | 0 | 100% |
| 3 | Osmanabad | 0 | 0 | 0 | 0 | 0 | 0 | 0 | 3 | 0 | 0 | 0 | 0 | 3 | 0 | 100% |
| 3 | Perbhani | 0 | 0 | 0 | 0 | 0 | 0 | 0 | 0 | 3 | 0 | 0 | 0 | 3 | 0 | 100% |
| 3 | Pune | 0 | 0 | 0 | 0 | 0 | 0 | 0 | 0 | 0 | 3 | 0 | 0 | 3 | 0 | 100% |
| 3 | Satara | 0 | 0 | 0 | 0 | 0 | 0 | 0 | 0 | 0 | 0 | 3 | 0 | 3 | 0 | 100% |
| 3 | Solapur | 0 | 0 | 0 | 0 | 0 | 0 | 0 | 0 | 0 | 0 | 0 | 3 | 3 | 0 | 100% |
| | | | | | | | | | | | | | Total | 36 | 0 | |
| | | | | | | | | | | | | | Final Result | 100% | | |

Table 3.b) Confusion Matrix for Acoustic Utterance Recognition using MFCC